\documentclass{article}
\PassOptionsToPackage{numbers, compress}{natbib}
\usepackage[preprint]{neurips_2026}
\usepackage[utf8]{inputenc} 
\usepackage[T1]{fontenc}    
\usepackage{hyperref}       
\usepackage{url}            
\usepackage{booktabs}       
\usepackage{amsfonts}       
\usepackage{nicefrac}       
\usepackage{microtype}      
\usepackage{xcolor}         
\usepackage[table]{xcolor}

\usepackage{graphicx}
\usepackage{amsmath}
\usepackage{multirow}
\usepackage{makecell}
\usepackage{algorithm}
\usepackage{algorithmic}

\usepackage{pifont}   
\usepackage{caption}  
\usepackage{booktabs}

\usepackage{booktabs}
\usepackage{amssymb}
\usepackage{pifont}

\newcommand{\cmark}{\textcolor{green}{\ding{51}}}
\newcommand{\xmark}{\textcolor{red}{\ding{55}}}

\title{Query-Conditioned Test-Time Self-Training for Large Language Models}
\author{Chaehee Song, Minseok Seo, Yeeun Seong, Doyi Kim, Changick Kim}
\author{%
    Chaehee Song\textsuperscript{*1} \quad
    Minseok Seo\textsuperscript{*1} \quad
    Yeeun Seong\textsuperscript{2} \quad
    Doyi Kim\textsuperscript{2} \quad
    Changick Kim\textsuperscript{$\dagger$1} \\[0.3em]
    \textsuperscript{1}School of Electrical Engineering, KAIST \\
    \textsuperscript{2}Graduate School of Green Growth and Sustainability, KAIST \\
    \texttt{\{chaehee.song, minseok.seo, yeeunseong, doyi.kim, changick\}@kaist.ac.kr} \\[0.3em]
    \textsuperscript{*}Equal contribution. \quad \textsuperscript{$\dagger$}Corresponding author.
  }

\begin{document}

\maketitle

\begin{abstract}
Large language models (LLMs) are typically deployed with fixed parameters, and their performance is often improved by allocating more computation at inference time.
While such test-time scaling can be effective, it cannot correct model misconceptions or adapt the model to the specific structure of an individual query.
Test-time optimization addresses this limitation by enabling parameter updates during inference, but existing approaches either rely on external data or optimize generic self-supervised objectives that lack query-specific alignment.
%
%
%
In this work, we propose Query-Conditioned Test-Time Self-Training (QueST), a framework that adapts model parameters during inference using supervision derived directly from the input query.
Our key insight is that the input query itself encodes latent signals sufficient for constructing structurally related problem--solution pairs.
Based on this, QueST generates such query-conditioned pairs and uses them as supervision for parameter-efficient fine-tuning at test time.
The adapted model is then used to produce the final answer, enabling query-specific adaptation without any external data.
Across seven mathematical reasoning benchmarks and the GPQA-Diamond scientific reasoning benchmark, QueST consistently outperforms strong test-time optimization baselines.
%
These results demonstrate that query-conditioned self-training is an effective and practical paradigm for test-time adaptation in LLMs.
Code is available at https://chssong.github.io/Query-Conditioned-TTST/.
\end{abstract}

\section{Introduction}
The success of large language models (LLMs)~\cite{yang2025qwen3, guo2025deepseek, ouyang2022training, stiennon2020learning, achiam2023gpt,touvron2023llama, iyer2022opt, wang2023self} is driven by massive data and large-scale computation. 
While continuously retraining and redeploying models is the most direct way to improve performance, it incurs prohibitive costs and is often impractical in real-world settings. 
As a practical alternative, performance is typically improved by increasing test-time computation~\cite{snell2024scaling, muennighoff2025s1, snell2025scaling} or refining prompts~\cite{wei2022chain, kojima2022large, zhou2022large}. 
However, user queries vary widely and demand different reasoning strategies, which makes adapting models to individual queries difficult through these approaches alone.

To address this limitation, recent work~\cite{zhou2026online, feng2026inplace, li2025test, hu2025test, hardt2023test} has explored test-time optimization (TTO), which adapts model parameters at inference time to better align with a given query.
Unlike test-time scaling, which leaves model parameters fixed, TTO modifies them directly, enabling more flexible and input-dependent behavior.

For example, TTT-NN~\cite{hardt2023test} improves performance by retrieving similar problems from an external database and fine-tuning the model on them. 
However, such approaches rely on large-scale data storage and retrieval, which introduces practical limitations.
In contrast, TLM~\cite{hu2025test} updates model parameters by minimizing the perplexity of the input itself, but this objective is not directly aligned with the specific reasoning structure required by individual queries.

Despite these limitations, two important observations emerge from prior work. 
First, model performance improves when it is adapted using problems that are structurally similar to the input query. 
Second, the input query itself contains latent signals that can guide effective adaptation.
These observations suggest that it is possible to construct meaningful supervision directly from the input query, without relying on external data, enabling more efficient test-time adaptation.

In this work, we propose Query-Conditioned Test-Time Self-Training (QueST), a framework that leverages this insight. 
Given a query, QueST treats it as a seed and generates structurally related problem--solution pairs conditioned on the input. 
These generated samples are then used to perform lightweight LoRA-based updates at test time. 
The adapted model is subsequently used to produce the final response, enabling dynamic and query-specific adaptation.

Beyond TTO, self-evolving methods~\cite{huang2025r, liu2025spice} also improve performance by generating and solving problems.
While QueST shares this generate-and-solve intuition, it differs fundamentally in approach.
Self-evolving methods typically rely on global data generation and carefully engineered prompts, whereas QueST performs localized adaptation directly conditioned on the input query, without requiring external data or complex prompt design.

These differences are summarized in Table~\ref{tab:method_comparison}.  Existing approaches satisfy only a subset of the properties required for effective test-time adaptation. 
For instance, test-time scaling and self-evolving methods do not perform per-query parameter updates at test time, and therefore cannot directly specialize model parameters to individual input queries.
While both TTT-NN and TLM leverage input-conditioned signals, TTT-NN relies on external data and TLM lacks explicit query-specific alignment.

In contrast, QueST is the only approach that satisfies all four key properties: query-specific adaptation, independence from external data, test-time parameter updates, and input-conditioned signal.
This enables QueST to provide a practical and effective framework for adapting large language models to individual queries at inference time.

We evaluate QueST on seven mathematical reasoning benchmarks, where it achieves state-of-the-art performance and improves average accuracy by 6.44 percentage points over the corresponding base models.
We further evaluate QueST on the GPQA-Diamond scientific reasoning benchmark, showing that QueST generalizes beyond mathematical reasoning.

\begin{table}[t!]
\centering
\setlength{\tabcolsep}{6pt}
\renewcommand{\arraystretch}{1.15}
\caption{Comparison of test-time improvement strategies for large language models across four key properties. While existing methods satisfy these properties only partially, QueST satisfies all of them, enabling effective and query-specific adaptation at inference time.}
\label{tab:method_comparison}
\vspace{2mm}

\resizebox{\linewidth}{!}{%
\begin{tabular}{lcccc}
\toprule
\textbf{Method}
& \makecell{\textbf{Query-specific}}
& \makecell{\textbf{External data-free}}
& \makecell{\textbf{Test-time update}}
& \makecell{\textbf{Input-conditioned signal}} \\
\midrule
\textbf{Test-time scaling} & \xmark & \cmark & \xmark & \xmark \\
\textbf{Self-evolving} & \xmark & \cmark & \xmark & \xmark \\ 
\textbf{TTT-NN} {\small\textcolor{gray}{(ICLR'24)}} & \cmark & \xmark & \cmark & \cmark \\
\textbf{TLM} {\small\textcolor{gray}{(ICML'25)}} & \xmark & \cmark & \cmark & \cmark \\
\textbf{QueST} & \cmark & \cmark & \cmark & \cmark \\
\bottomrule
\end{tabular}%
}
\end{table}

\begin{figure*}[t!]
    \centering
    \scalebox{1.00}{
    \includegraphics[width=1.0\textwidth]{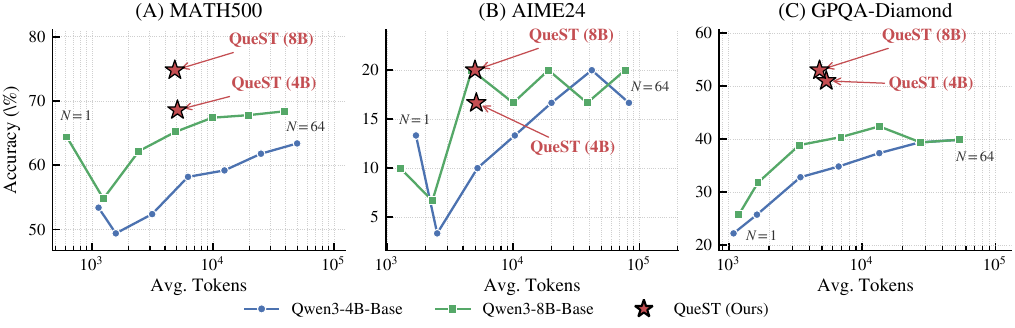}}
    \caption{Comparison of token usage and performance across three benchmarks: (a) MATH500 (high-school level mathematics), (b) AIME24 (advanced mathematical reasoning), and (c) GPQA-Diamond (general scientific reasoning). Each point on the baseline curves corresponds to sampling-based TTS using self-consistency decoding with $N \in \{1, 2, 4, 8, 16, 32, 64\}$ samples. Star markers denote QueST. Across all three benchmarks, QueST achieves superior accuracy while using substantially fewer test-time tokens.}
    \label{fig:comp} 
\end{figure*}

\section{Related work}
\subsection{Test-time Scaling}
Test-time scaling (TTS)~\cite{muennighoff2025s1, snell2024scaling} improves model performance by allocating additional computation during inference.
Early approaches are rooted in Chain-of-Thought (CoT) prompting~\cite{wei2022chain}, which enables step-by-step reasoning at inference time.
Building on this, subsequent methods improve performance by sampling multiple reasoning trajectories~\cite{wang2025sampling} and aggregating them via self-consistency~\cite{wang2022self}, or by selecting the best candidate from multiple outputs using Best-of-N strategies~\cite{cobbe2021training}.
More recent approaches further enhance TTS by incorporating outcome verifiers and process reward models (PRMs)~\cite{lightman2023let}, which evaluate final answers or intermediate reasoning steps, often combined with search strategies such as beam search or tree search~\cite{yao2023tree}.
In addition, iterative refinement methods, such as Self-Refine~\cite{madaan2023self} and Reflexion~\cite{shinn2023reflexion}, improve performance through repeated self-correction and feedback.

Despite these advances, TTS methods operate without updating model parameters, relying solely on increasingly sophisticated sampling, search, and self-correction procedures at inference time.
As a result, their effectiveness is fundamentally bounded by the capabilities already encoded in the base model, limiting their ability to correct underlying misconceptions or directly adapt model parameters to the specific structure of individual queries.
Moreover, TTS methods incur substantial token costs. As shown in Figure~\ref{fig:comp}, QueST achieves higher performance than sampling-based TTS baselines while requiring fewer tokens.
\subsection{Self-evolving}
Self-evolving methods aim to reduce reliance on expensive human-curated data by enabling models to generate tasks and improve through accumulated experience.
This line of work is inspired by self-play in reinforcement learning, where AlphaGo~\cite{silver2016mastering} combined human demonstrations with self-play, while AlphaZero~\cite{silver2018general} achieved superhuman performance without human-annotated training data.

Recent efforts have extended similar ideas in LLMs. 
R-Zero~\cite{huang2025r} proposes a co-evolution framework in which challenger and solver models generate and learn from their own tasks, while SPICE~\cite{liu2025spice} extends this paradigm through corpus-grounded self-play using external document collections.
%
Despite its effectiveness, SPICE remains tied to human-produced corpora, indicating that fully self-driven improvement in LLMs is still an open direction.

Unlike board games, natural language tasks often lack a well-defined environment or reliable automatic verification signals.
As a result, self-evolving LLMs often face fundamental challenges in maintaining data quality, training stability, and consistent learning dynamics.

While QueST shares the intuition of generating and solving tasks for self-improvement, it fundamentally differs from prior self-evolving approaches.
Existing self-evolving methods operate in an offline setting, where models iteratively accumulate and learn from generated data.
In contrast, QueST performs \emph{online, query-conditioned adaptation} by generating task-specific QA pairs directly from a given user query at test time, enabling immediate and targeted model adaptation.

\subsection{Test-time Optimization}
Test-time optimization (TTO)~\cite{wang2020tent, liang2025comprehensive, seo2025upsample} updates model parameters during inference using signals available at test time, thereby specializing the model to the test input or target distribution.
This paradigm has been extensively studied in computer vision, where performance degradation is often attributed to input distribution shifts.
Accordingly, most methods adapt a pretrained model by exploiting auxiliary signals available at inference time.

A representative approach in vision is TENT~\cite{wang2020tent}, which updates model parameters by minimizing prediction entropy at test time, typically by adapting only normalization layers.

Recently, TTO has been extended to large language models.
For example, TTT-NN~\cite{hardt2023test} retrieves similar samples from a large external corpus conditioned on the input query and fine-tunes the model before prediction, making its adaptation input-conditioned but dependent on external data.
TLM~\cite{hu2025test} adapts LLMs using unlabeled test data via input perplexity minimization, and is primarily designed for domain-level adaptation under distribution shift, rather than explicitly targeting query-specific optimization.

In contrast, QueST performs test-time optimization using supervision generated directly from the input query, achieving both query-specific adaptation and independence from external data, two properties that prior TTO methods satisfy only partially.
Moreover, QueST enables efficient and targeted adaptation at the level of individual queries through parameter-efficient updates. 

\begin{figure*}[t!]
    \centering
    \scalebox{1.00}{
    \includegraphics[width=1.0\textwidth]{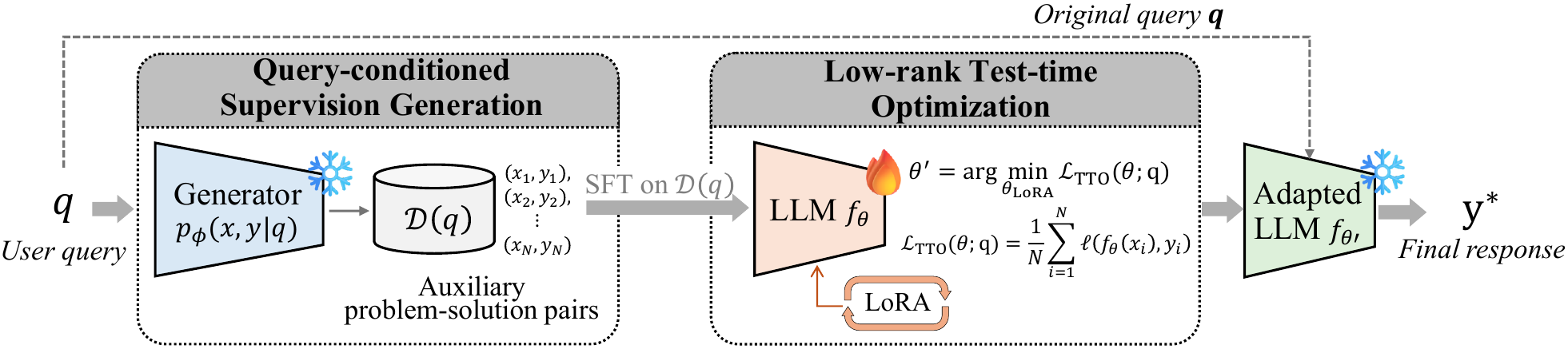}}
    \caption{Overview of QueST. Given a user query $q$, QueST first generates query-conditioned auxiliary problem--solution pairs $\mathcal{D}(q)$ that reflect the underlying reasoning patterns of the query. These generated samples serve as supervision for low-rank test-time optimization via LoRA, enabling efficient parameter adaptation at inference time. The adapted model then produces the final response $y^*$, achieving query-specific adaptation without relying on external data.}
    \label{fig:main} 
\end{figure*}

\section{Method}
\subsection{Problem Definition}
Given a pre-trained language model $f_\theta$ and a user query $q$, the goal is to generate an accurate response $y$. In standard inference, the model directly predicts $y \sim p_\theta(y \mid q)$ without any adaptation.

However, real-world queries often follow a distribution that differs from the pre-training distribution, leading to suboptimal performance on domain-specific or challenging inputs. Ideally, we would like to adapt the model to the local distribution induced by the query $q$.

Test-time optimization (TTO) addresses this by updating model parameters at inference time, and then using the adapted parameters $\theta'$ to generate the final response:
\begin{equation}
\theta' = \theta - \eta \nabla_\theta \mathcal{L}_{\text{TTO}}(\theta; q).
\end{equation}


The key challenge lies in defining an informative and reliable test-time objective $\mathcal{L}_{\text{TTO}}$, since natural language queries do not provide explicit labels or structured constraints. Existing approaches either rely on external data or optimize generic self-supervised objectives, both of which provide weak or misaligned supervision for the reasoning structure of the current query.

QueST addresses this by constructing query-conditioned supervision $\mathcal{D}(q)$ from the input query and using it to define the test-time objective $\mathcal{L}_{\text{TTO}}$.

\subsection{Query-conditioned Supervision Generation}

We assume that $q$ is sampled from an underlying local data distribution $\mathcal{D}_q$, 
which consists of problems sharing similar concepts, assumptions, and reasoning patterns. 
Under this assumption, solving $q$ can be viewed as generalizing within $\mathcal{D}_q$. 
Therefore, optimizing the model on samples that approximate $\mathcal{D}_q$ can improve its ability to solve the original query $q$.

Existing approaches either retrieve samples from an external dataset $\mathcal{D}_{\text{ext}}$, or generate pseudo-labels using the model itself:
\begin{equation}
(x_i, y_i) \sim p_\theta(x, y),
\end{equation}
both of which are independent of the specific query $q$ and often lead to weak or misaligned supervision.

In contrast, QueST generates a dataset $\mathcal{D}(q)$ consisting of auxiliary problem--solution pairs conditioned on $q$:
\begin{equation}
\mathcal{D}(q) = \{(x_i, y_i)\}_{i=1}^{N}, \quad (x_i, y_i) \sim p_\phi(x, y \mid q),
\end{equation}
where $p_\phi$ denotes the auxiliary problem generator. In practice, we use the base model itself as the generator.
Each generated pair shares core concepts and reasoning patterns with the original query, so that $\mathcal{D}(q)$ serves as an empirical approximation of $\mathcal{D}_q$ and provides query-aligned supervision.

Using this dataset, we define the test-time objective as
\begin{equation}
\mathcal{L}_{\text{TTO}}(\theta; q) = \frac{1}{N} \sum_{i=1}^{N} \ell(f_\theta(x_i), y_i),
\end{equation}
where $\ell$ denotes the cross-entropy loss; minimizing this objective adapts the model to the local distribution around $q$.

\subsection{Low-rank Test-time Optimization}

To adapt the model efficiently at test time, we employ parameter-efficient fine-tuning using low-rank adaptation (LoRA)~\cite{hu2022lora}. 
Full fine-tuning is computationally expensive and unstable in a test-time setting, especially when only a small number of supervision samples are available. 
Therefore, we restrict updates to a low-rank parameterization.

Given the query-conditioned dataset $\mathcal{D}(q)$, we optimize the model parameters by minimizing the following objective:
\begin{equation}
\theta' = \arg\min_{\theta_{\text{LoRA}}} \mathcal{L}_{\text{TTO}}(\theta; q),
\end{equation}
where only $\theta_{\text{LoRA}}$ is updated and $\theta$ remains frozen, yielding the adapted parameters $\theta'$.

We perform $T$ optimization steps using standard supervised fine-tuning (SFT) on $\mathcal{D}(q)$. 
This lightweight adaptation is sufficient to capture the local structure of the query-conditioned distribution while maintaining computational efficiency and stability.
After optimization, the adapted model $f_{\theta'}$ generates the final response $y^* \sim p_{\theta'}(y \mid q)$ for the original query $q$. 

An overview of the QueST framework is illustrated in Figure~\ref{fig:main}, and the complete procedure is summarized in Algorithm~\ref{alg:QueST}.

\begin{algorithm}[t]
\caption{Query-Conditioned Test-Time Self-Training (QueST)}
\label{alg:QueST}
\small
\begin{algorithmic}[1]
\REQUIRE Pre-trained model $f_\theta$, query $q$
\STATE Generate query-conditioned dataset $\mathcal{D}(q) = \{(x_i, y_i)\}_{i=1}^{N}$
\STATE Initialize LoRA parameters $\theta_{\text{LoRA}}$
\FOR{$t = 1$ to $T$}
    \STATE Sample $(x_i, y_i) \sim \mathcal{D}(q)$
    \STATE Update $\theta_{\text{LoRA}}$ by minimizing $\ell(f_\theta(x_i), y_i)$
\ENDFOR
\STATE Obtain adapted model $f_{\theta'}$
\STATE Generate final answer $y^* \sim p_{\theta'}(y \mid q)$
\RETURN $y^*$
\end{algorithmic}
\end{algorithm}

\newcommand{\eqcolW}{\dimexpr (\textwidth - 2.7cm - 16\tabcolsep)/8\relax}
\begin{table*}[t!]
\centering
\small
\setlength{\tabcolsep}{3pt}
\renewcommand{\arraystretch}{1.2}
\caption{Quantitative comparison of QueST with existing test-time optimization methods across seven mathematical benchmarks.
All experiments are conducted under the same experimental setup, following the official implementations of each method.}
\begin{tabular}{l*{8}{>{\centering\arraybackslash}p{\eqcolW}}}
\toprule
\textbf{Model Name} & \textbf{Average} & \textbf{AMC} & \textbf{Minerva} & \textbf{MATH} & \textbf{GSM8K} & \textbf{Olympiad} & \textbf{AIME25} & \textbf{AIME24} \\
\midrule

\multicolumn{9}{l}{\textit{Qwen3-4B}} \\
Base Model & 36.59 & 45.00 & 21.32 & 55.20 & 91.05 & 23.56 & 6.67 & 13.33 \\
TENT & 36.69 & 47.50 & 21.69 & 54.20 & 91.05 & 22.37 & 10.00 & 10.00 \\
TLM & 36.77 & 45.00 & 19.49 & 53.60 & 90.67 & 21.93 & 6.67 & 20.00 \\
\rowcolor{gray!15} \textbf{QueST (ours)} & \textbf{42.58} & \textbf{52.50} & \textbf{23.10} & \textbf{70.80} & \textbf{92.10} & \textbf{26.20} & \textbf{13.33} & \textbf{20.00} \\
\midrule

\multicolumn{9}{l}{\textit{Qwen3-8B}} \\
Base Model & 24.66 & 27.50 & 13.24 & 42.60 & 75.66 & 13.63 & 0.00 & 0.00 \\
TENT & 23.86 & 25.00 & 14.71 & 40.20 & 74.37 & 12.74 & 0.00 & 0.00 \\
TLM & 26.28 & 32.50 & 12.13 & 37.40 & 74.98 & 13.63 & 6.67 & 6.67 \\
\rowcolor{gray!15} \textbf{QueST (ours)} & \textbf{31.44} & \textbf{35.00} & \textbf{15.80} & \textbf{54.60} & \textbf{78.20} & \textbf{16.50} & \textbf{6.67} & \textbf{13.33} \\
\midrule

\multicolumn{9}{l}{\textit{Qwen3-4B-Base}} \\
Base Model & 34.53 & 35.00 & 22.06 & 53.40 & 90.14 & 24.44 & 3.33 & 13.33 \\
TENT & 36.61 & 47.50 & 23.16 & 53.00 & 88.78 & 23.85 & 10.00 & 10.00 \\
TLM & 36.20 & 42.50 & 22.43 & 54.60 & 89.31 & 24.59 & 10.00 & 10.00 \\
\rowcolor{gray!15} \textbf{QueST (ours)} & \textbf{41.74} & \textbf{50.00} & \textbf{24.10} & \textbf{68.60} & \textbf{91.66} & \textbf{27.80} & \textbf{13.33} & \textbf{16.66} \\\midrule

\multicolumn{9}{l}{\textit{Qwen3-8B-Base}} \\
Base Model & 42.00 & 57.50 & 26.84 & 64.40 & 92.65 & 32.59 & 10.00 & 10.00 \\
TENT & 41.59 & 52.50 & 26.47 & 65.00 & 92.27 & 31.56 & 13.33 & 10.00 \\
TLM & 42.58 & 62.50 & 25.74 & 65.40 & 91.81 & 32.59 & 10.00 & 10.00 \\
\rowcolor{gray!15} \textbf{QueST (ours)} & \textbf{47.77} & \textbf{65.00} & \textbf{28.90} & \textbf{74.80} & \textbf{93.20} & \textbf{35.80} & \textbf{16.67} & \textbf{20.00} \\
\midrule

\multicolumn{9}{l}{\textit{OctoThinker-3B}} \\
Base Model & 21.74 & 25.00 & 11.76 & 36.60 & 60.27 & 11.85 & 0.00 & 6.67 \\
TENT & 20.14 & 17.50 & 11.40 & 37.60 & 59.59 & 11.56 & 0.00 & 3.33 \\
TLM & 22.04 & 20.00 & 12.50 & 38.80 & 60.80 & 12.15 & 3.33 & 6.67 \\
\rowcolor{gray!15} \textbf{QueST (ours)} & \textbf{28.59} & \textbf{27.50} & \textbf{19.12} & \textbf{46.40} & \textbf{67.85} & \textbf{19.26} & \textbf{10.00} & \textbf{10.00} \\\midrule

\multicolumn{9}{l}{\textit{OctoThinker-8B}} \\
Base Model & 28.05 & 25.00 & 13.97 & 45.80 & 79.91 & 18.37 & 3.33 & 10.00 \\
TENT & 27.32 & 25.00 & 13.60 & 45.20 & 79.38 & 18.07 & 3.33 & 6.67 \\
TLM & 28.76 & 27.50 & 14.71 & 46.60 & 80.52 & 18.67 & 3.33 & 10.00 \\
\rowcolor{gray!15} \textbf{QueST (ours)} & \textbf{34.11} & \textbf{35.00} & \textbf{19.49} & \textbf{52.00} & \textbf{84.53} & \textbf{24.44} & \textbf{10.00} & \textbf{13.33} \\\bottomrule
\end{tabular}
\label{tab:main}
\end{table*}

\section{Experiments}
\label{sec:experiments}

\subsection{Experimental Setup}

We evaluate QueST to assess whether query-conditioned supervision provides stronger adaptation signals than the alternative test-time objectives used by prior test-time optimization methods.
Our comparisons include TENT~\cite{wang2020tent}, a widely used test-time adaptation method originally developed for computer vision, and TLM~\cite{hu2025test}, a recent test-time learning approach for large language models.

We conduct experiments on multiple model architectures to demonstrate the generality of our method. 
Specifically, we use Qwen3-4B-Base and Qwen3-8B-Base as representative base LLMs, Qwen3-4B and Qwen3-8B as the corresponding post-trained variants, and OctoThinker-3B and OctoThinker-8B~\cite{wang2025octothinker} as reasoning-oriented models.
For post-trained Qwen3 models, all evaluations are performed in \texttt{non-thinking} mode, which is the model's default inference mode without explicit reasoning chains, to isolate the effect of QueST from the model's built-in reasoning traces.

We evaluate on a diverse set of benchmarks covering both mathematical and scientific reasoning. 
For mathematical reasoning, we use seven datasets: AMC~\cite{maa_amc}, Minerva~\cite{lewkowycz2022solving}, MATH500~\cite{hendrycks2021measuring, lightman2023let}, GSM8K~\cite{cobbe2021training}, OlympiadBench~\cite{he2024olympiadbench}, AIME25~\cite{maa_aime}, and AIME24~\cite{maa_aime}. 
To assess generality beyond mathematical reasoning, we additionally evaluate on GPQA-Diamond~\cite{rein2023gpqa}, a challenging benchmark for graduate-level scientific reasoning.

\subsection{Implementation Details}
\label{sec:details}

All experiments are conducted on a single NVIDIA H200 GPU. 
For QueST, we use LoRA~\cite{hu2022lora} with rank 16 and perform $T$ optimization steps using AdamW with a learning rate of $1\times10^{-4}$.
We set $T = 10$ for most benchmarks and $T = 40$ for AIME24 and AIME25, which require longer reasoning.
For each query, we generate $N{=}5$ auxiliary problem--solution pairs for test-time adaptation.
The optimization objective is the standard SFT loss, i.e., cross-entropy between the model predictions and the corresponding generated solution tokens for each auxiliary problem.

For fair comparison, all methods use the same system prompt:
\textit{``Please reason step by step, and put your final answer within $\backslash$boxed\{\}. Carefully reconsider the underlying concept of the question.''}
Additional implementation details are provided in Appendix~\ref{sec:details_supple}.

\begin{figure*}[t!]
    \centering
    \scalebox{1.00}{
    \includegraphics[width=1.0\textwidth]{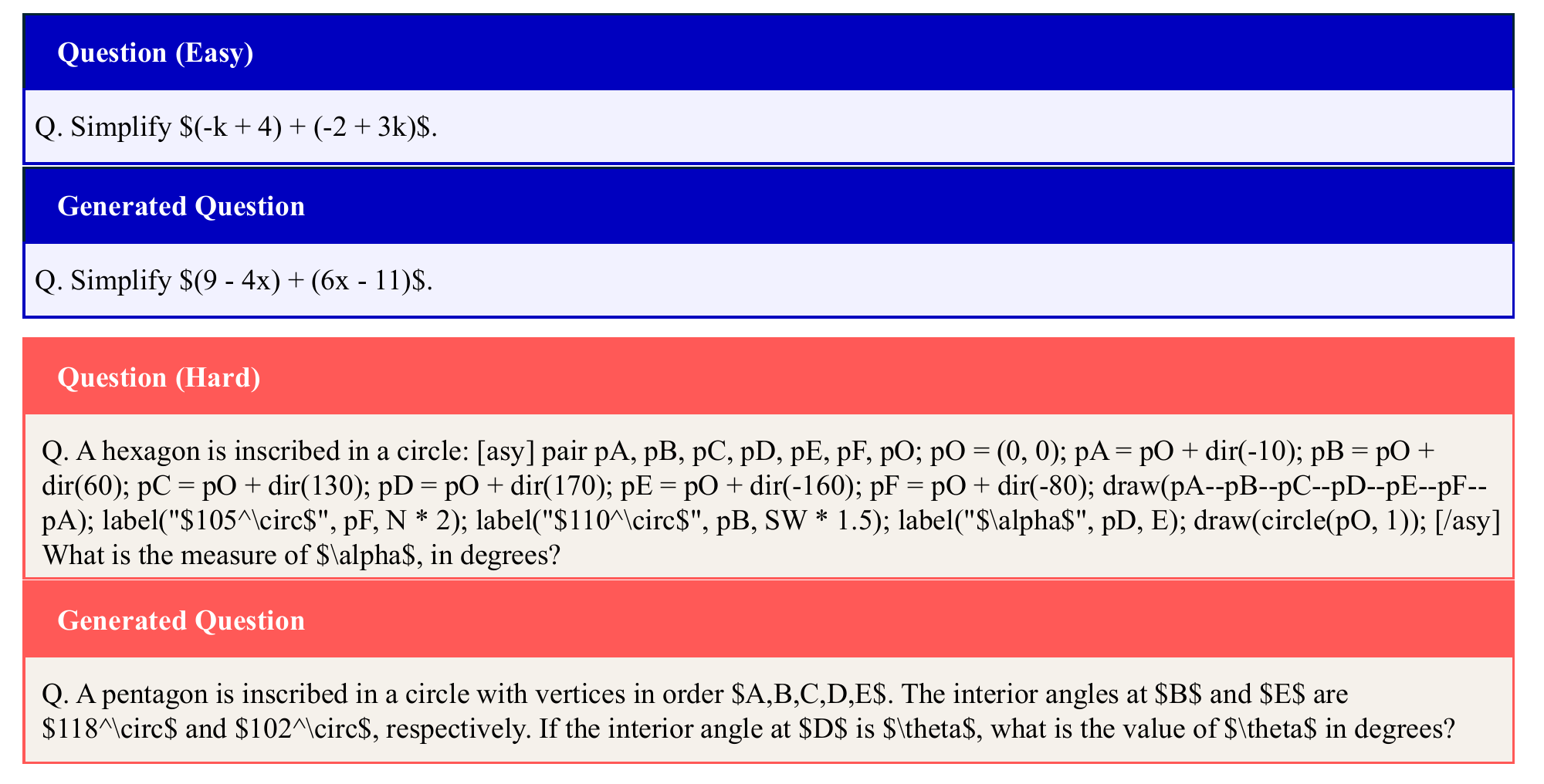}}
    \caption{Qualitative examples of problems generated by QueST from input queries. The top row shows a simple surface-level variation, while the bottom row presents a more challenging structural transformation, demonstrating the range of generated tasks.}
    \label{fig:sample1} 
\end{figure*}

\subsection{Mathematical Benchmark Results}

Table~\ref{tab:main} presents the performance of QueST across seven mathematical benchmarks, compared with existing TTO methods.
We exclude TTT-NN~\cite{hardt2023test} from comparison, as it relies on retrieving similar samples from a large external corpus, which cannot be fairly reproduced under our experimental setup due to the lack of access to the original retrieval database.

Across most datasets, prior TTO methods such as TENT and TLM show only marginal improvements, or even performance degradation.
This suggests that entropy minimization and perplexity reduction provide weak or misaligned signals for structured, multi-step mathematical reasoning.

In contrast, QueST consistently achieves the largest performance gains across all benchmarks and model configurations, with average improvements of +5.77 to +7.21 percentage points over the base models, establishing state-of-the-art results.
This demonstrates that combining query-conditioned supervision with test-time optimization is particularly effective for reasoning-intensive tasks, where adapting the model to the specific problem structure is critical.

\subsection{Qualitative Analysis}

\begin{figure*}[t!]
    \centering
    \scalebox{1.00}{
    \includegraphics[width=1.0\textwidth]{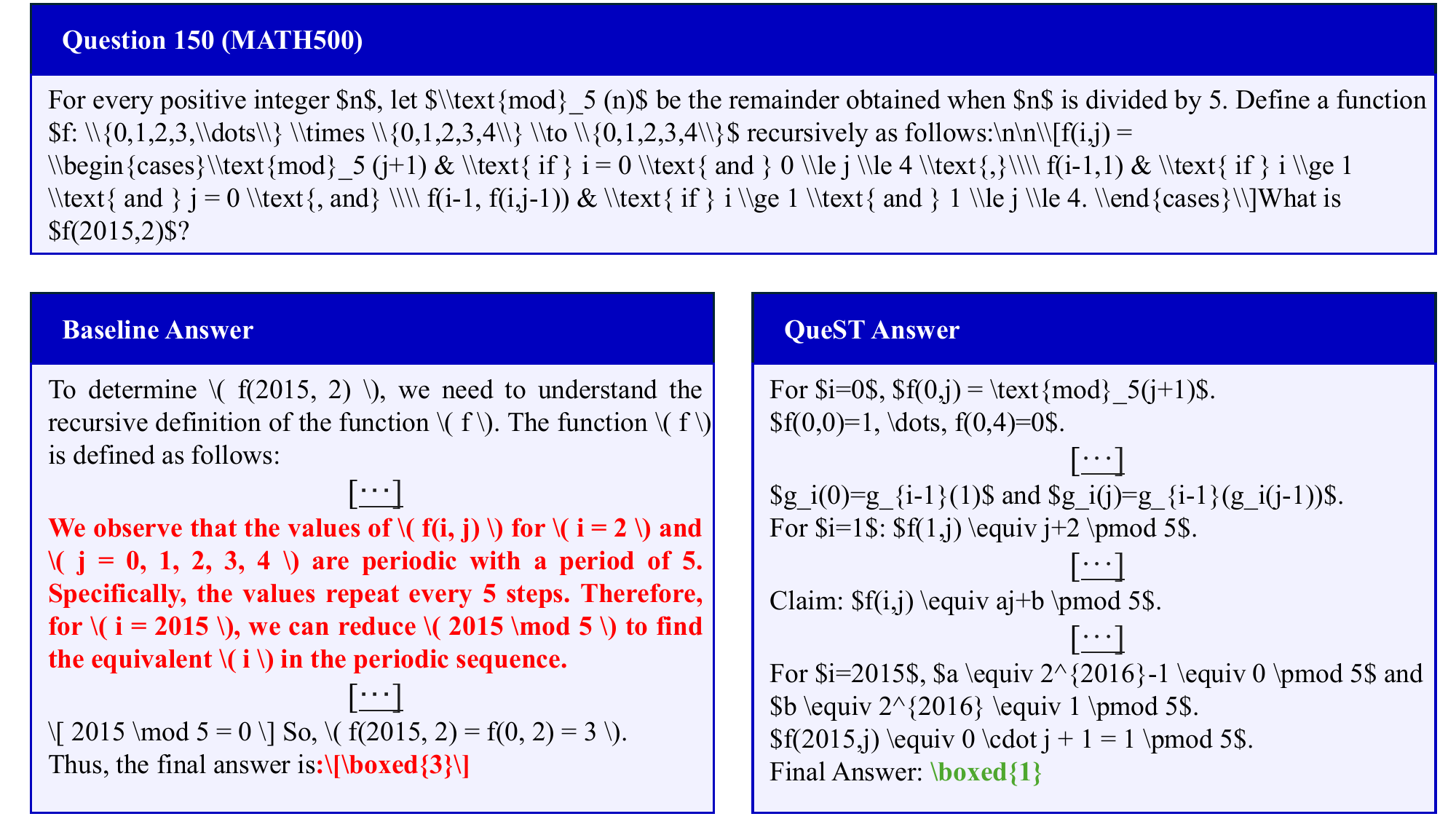}}
    \caption{An example demonstrating the effectiveness of QueST in correcting erroneous predictions. While the baseline model fails to produce the correct answer, QueST adapts at test time using query-conditioned supervision and successfully recovers the correct solution.}
    \label{fig:sample2} 
\end{figure*}
\paragraph{Query-conditioned Problem Generation.}
Figure~\ref{fig:sample1} presents qualitative examples of problems generated by QueST from input queries, with varying levels of transformation.
The top example preserves the algebraic structure of the original question, with only minor changes in coefficients and variables.
In contrast, the bottom example reformulates a diagram-based hexagon problem into a symbolic pentagon problem with different reasoning requirements.

These examples indicate that QueST can generate problems ranging from surface-level variations to structural transformations depending on the input query.
Consistent with prior work showing the diversity of self-generated problems~\cite{zhao2025absolute, huang2025r, wang2023self}, QueST further leverages these generated problems as supervision signals for test-time adaptation.
\paragraph{Error Correction.}
Figure~\ref{fig:sample2} presents a case where QueST corrects an incorrect prediction by leveraging problems generated from the user query.
The base model fails to reflect the recursive structure of the function and instead relies on a simplified periodicity assumption, leading to an incorrect answer.
This reflects a common failure mode where language models depend on superficial patterns rather than underlying structure.

In contrast, QueST generates additional problems from the user query and uses them as auxiliary information at test time.
By considering multiple input--output examples with similar structure, the model's prediction shifts toward patterns more consistent with the problem, ultimately producing the correct answer.
This example shows that QueST does more than refine output formatting; it can alter the model's test-time behavior in a way that better aligns with the structure of the query.

\subsection{Ablation Study}
Table~\ref{tab:ablation} presents ablation and cross-domain analysis for QueST.

\paragraph{Component Contributions.} 
The left table evaluates the contribution of query conditioning, LoRA adaptation, and self-generated QA.
As shown, self-generated QA without query conditioning yields limited performance, and adding LoRA alone provides only marginal improvement.
In contrast, query-conditioned QA substantially improves accuracy, with the best result obtained when combined with LoRA.
This indicates that query-specific supervision is critical for effective test-time adaptation.

\paragraph{Number of Generated Problems.}
The middle table analyzes the effect of the number of generated problems.
Performance consistently improves as the number of generated problems increases, with the best result obtained using $N=5$.
This suggests that additional query-conditioned supervision provides more useful adaptation signals for test-time optimization. 
Thus, we use $N=5$ in the main experiments, balancing performance and computational cost.
Extended ablation with larger $N$ is provided in Appendix~\ref{sec:extended_n}.

\paragraph{Cross-domain Generalization.}
We evaluate QueST on the scientific reasoning benchmark GPQA-Diamond.
QueST consistently improves performance across both base and post-trained Qwen3-4B models.
These results suggest that QueST generalizes beyond mathematical reasoning tasks and is effective across different domains.
\begin{table}[t]
\centering
\small
\setlength{\tabcolsep}{4pt}
\renewcommand{\arraystretch}{1.1}
\caption{Ablation and analysis of QueST. (Left) Contribution of query conditioning, LoRA adaptation, and self-generated QA on Qwen3-4B-Base using MATH500. (Middle) Effect of the number of generated problems used during test-time optimization on Qwen3-4B-Base, evaluated on MATH500. (Right) QueST applied to Qwen3-4B-Base and Qwen3-4B on GPQA-Diamond.}
\vspace{2mm}
\label{tab:ablation}

\begin{minipage}[t]{0.32\linewidth}
\centering
\begin{tabular}{ccc|c}
\toprule
\textbf{Query} & \textbf{LoRA} & \textbf{Self-QA} & \textbf{Acc} \\
\midrule
$\times$ & $\times$ & $\checkmark$ & 58.00 \\
$\times$ & $\checkmark$ & $\checkmark$ & 58.80 \\
$\checkmark$ & $\times$ & $\checkmark$ & 66.60 \\
$\checkmark$ & $\checkmark$ & $\checkmark$ & \textbf{68.60} \\
\bottomrule
\end{tabular}
\end{minipage}
\hfill
\begin{minipage}[t]{0.18\linewidth}
\centering
\begin{tabular}{c|c}
\toprule
\#\textbf{Problems} & \textbf{Acc} \\
\midrule
1 & 57.80 \\
3 & 63.00 \\
4 & 64.20 \\
5 & \textbf{68.60} \\
\bottomrule
\end{tabular}
\end{minipage}
\hfill
\begin{minipage}[t]{0.42\linewidth}
\centering
\begin{tabular}{l|c}
\toprule
\textbf{Model} & \textbf{Acc} \\
\midrule
Qwen3-4B-Base & 22.22 \\
Qwen3-4B-Base + QueST & \textbf{46.97} \\ \hline
Qwen3-4B & 32.32 \\
Qwen3-4B + QueST & \textbf{51.01} \\
\bottomrule
\end{tabular}
\end{minipage}

\end{table}

\section{Limitations and Future Work}
\label{sec:limitations}

QueST relies on supervision signals derived from user queries, and its effectiveness depends on the quality and informativeness of the input.
When queries are ambiguous, underspecified, or contain incorrect assumptions, the generated supervision may be weak or misleading, which can limit the effectiveness of test-time adaptation.

As future work, one direction is to move beyond per-query adaptation and explore persistent test-time optimization (continual learning).
In the current framework, model parameters are reset for each query, preventing knowledge accumulation over time.
Allowing the model to update parameters across queries could enable incremental specialization to recurring patterns and improve performance in specific domains.

However, this direction introduces challenges such as error accumulation and model drift, especially under noisy supervision.
Despite these challenges, the consistent improvements achieved by QueST suggest that query-driven adaptation is a promising foundation for progressively adaptive models.

\paragraph{Broader Impact.}
QueST moves LLMs from static inference toward adaptive inference, loosely resembling how humans refine their reasoning during a test by using the current problem to infer related patterns. This may enable more flexible and cost-efficient AI systems without external data or full retraining. At the same time, as in human problem solving, adaptation can fail or even reinforce mistakes when the query is ambiguous, misleading, or based on false premises, requiring safeguards against unreliable supervision and error accumulation.
%
%

\section{Conclusion}
In this work, we propose Query-Conditioned Test-Time Self-Training (QueST), a framework grounded in the observation that user queries themselves contain rich implicit signals for test-time adaptation. 
Instead of relying on external data or purely self-supervised signals, QueST constructs supervision directly from the input query by generating auxiliary problem--solution pairs, and performs efficient parameter adaptation via LoRA at test time.
Across seven mathematical reasoning benchmarks and the GPQA-Diamond scientific reasoning benchmark, QueST consistently outperforms base models and existing test-time optimization baselines based on entropy or perplexity minimization, achieving state-of-the-art performance.
%
%
Beyond performance, QueST highlights a new perspective on test-time optimization: rather than merely adapting representations, it enables models to actively acquire useful knowledge during inference. 
We hope this work opens up new directions for test-time learning in large language models, particularly toward more robust and knowledge-driven adaptation.


\bibliographystyle{plainnat}
\bibliography{neurips_2026}

\newpage
\appendix



\section{Implementation Details}
\label{sec:details_supple}
\subsection{Detailed Hyperparameters}
Table~\ref{tab:hyperparams} reports the complete set of hyperparameters used in our QueST experiments. We use the same configuration across all six model architectures (Qwen3-4B-Base, Qwen3-8B-Base, Qwen3-4B, Qwen3-8B, OctoThinker-3B, OctoThinker-8B) without per-model tuning.

\begin{table}[H]
\centering
\small
\setlength{\tabcolsep}{8pt}
\renewcommand{\arraystretch}{1.15}
\caption{Detailed hyperparameters of QueST.}
\label{tab:hyperparams}
\vspace{2mm} 
\begin{tabular}{ll}
\toprule
\textbf{Hyperparameter} & \textbf{Value} \\
\midrule
\multicolumn{2}{l}{\textit{LoRA configuration}} \\
\midrule
Rank ($r$)                          & 16 \\
Alpha ($\alpha$)                    & 32 \\
Dropout                             & 0.05 \\
Target modules                      & \texttt{q\_proj}, \texttt{k\_proj}, \texttt{v\_proj},
\texttt{o\_proj} \\
Bias                                & none \\
\midrule
\multicolumn{2}{l}{\textit{Test-time optimization}} \\
\midrule
Optimizer                           & AdamW \\
Learning rate                       & $1 \times 10^{-4}$ \\
Number of optimization steps ($T$)  & 10 (default) / 40 (AIME24, AIME25) \\
Batch size                          & 1 \\
Gradient accumulation steps         & 4 \\
Effective batch size                & 4 \\
Max sequence length                 & 4096 \\
Precision                           & bf16 \\
Loss                                & cross-entropy (next-token prediction) \\
\midrule
\multicolumn{2}{l}{\textit{Auxiliary problem generation}} \\
\midrule
Number of pairs ($N$)               & 5 \\
Max new tokens                      & 4096 \\
Generator model $p_\phi$            & same as base model $f_\theta$ \\
\midrule
\multicolumn{2}{l}{\textit{Final response inference}} \\
\midrule
Decoding                            & greedy (\texttt{do\_sample=False}) \\
Max new tokens                      & 4096 \\
\bottomrule
\end{tabular}
\end{table}

\begin{figure*}[t!]
    \centering
    \scalebox{1.00}{
    \includegraphics[width=1.0\textwidth]{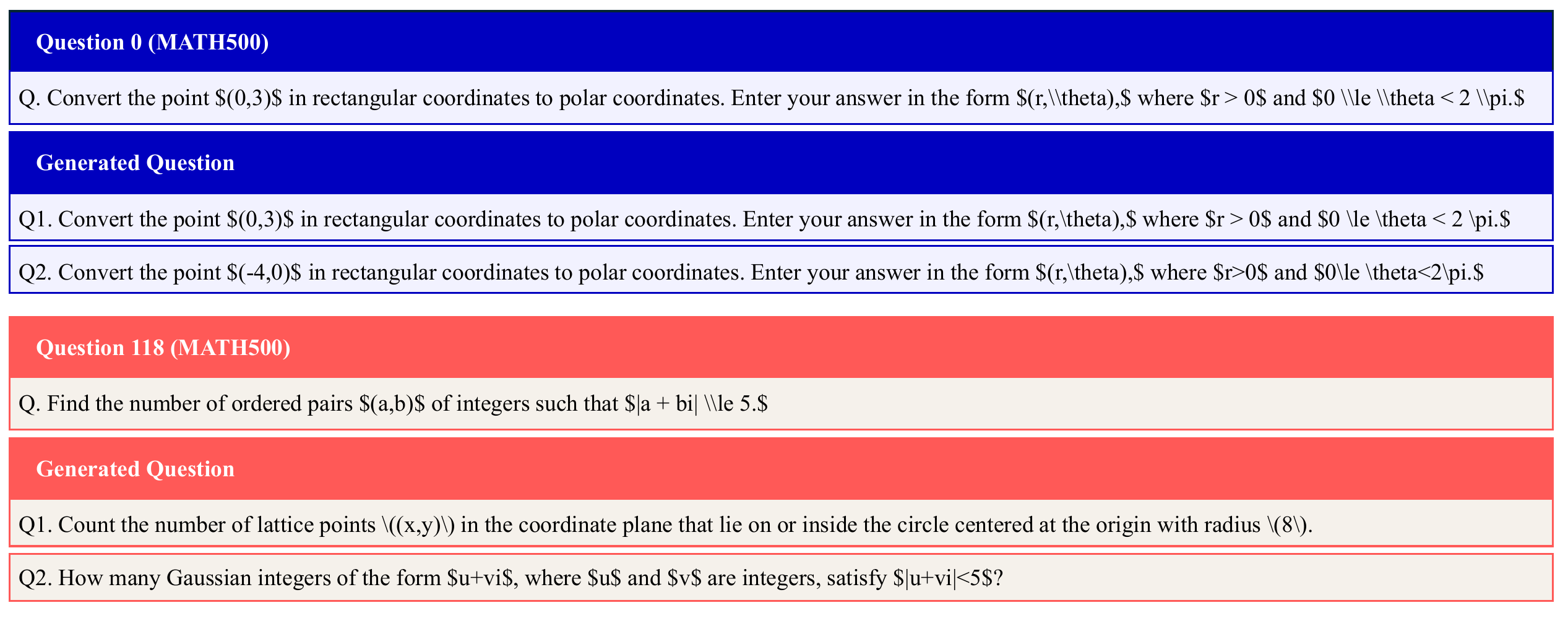}}
    \caption{Comparison of auxiliary supervision generated by QueST for two MATH500 queries under the same generator, prompt, and sampling temperature. Top (Question~$0$): all variants stay within a single canonical template, and one is a near-duplicate of the original query. Bottom (Question~$118$): the variants reframe the same lattice-point task in two different mathematical languages: a Cartesian circle and Gaussian integers.}
    \label{fig:sample5} 
\end{figure*}

\section{Qualitative Analysis of Query-adaptive Supervision}
QueST applies the same generator, prompt, and sampling temperature to every input query.
We examine whether, under this fixed configuration, the generated auxiliary supervision reflects the mathematical structure of the input query.
We illustrate this with two MATH500 examples that differ in how easily they admit equivalent reformulations.
The first is a coordinate-conversion query, whose generated variants remain within a single canonical template.
The second is a complex-modulus query that can be reformulated as lattice-point counting in a Euclidean disk, allowing the generated variants to span multiple mathematical descriptions.

\paragraph{Template-bound Supervision.}
The first query asks for the polar coordinates of $(0,3)$.
The reasoning required is a two-step lookup: read $r=|y|$ and assign $\theta$ from the axis on which the point lies.
The auxiliary problems QueST generates remain entirely within this template (Figure~\ref{fig:sample5}, top).
One variant is a near-duplicate of the original query, while the other substitutes a different axis-aligned point.
Each variant requires identical reasoning steps, and the supervision is highly redundant.

\paragraph{Framework-spanning Supervision.}
The second query asks for the number of integer pairs $(a,b)$ satisfying $|a+bi|\le 5$, framed in the language of complex moduli.
The underlying mathematical task is counting lattice points within a Euclidean disk, which admits multiple equivalent framings, and the auxiliary problems QueST generates span these framings (Figure~\ref{fig:sample5}, bottom).
One variant restates the task in Cartesian terms (lattice points inside a circle of a given radius), and another invokes the Gaussian-integer framework explicitly.
Both variants require the same underlying counting argument, yet expose the adapted model to qualitatively different mathematical vocabularies and solution narratives.

\paragraph{Interpretation.}
Holding the generator fixed, the framing diversity of the auxiliary supervision tracks the structural richness of the input query.
For a query whose reasoning admits a single canonical template, the supervision collapses onto that template and occasionally produces near-duplicates of the query itself.
For a query whose underlying task admits multiple equivalent framings, the supervision spans those framings naturally, exposing the adapted model to several mathematical languages of the same problem.
This provides qualitative evidence that the input query encodes a latent signal not only about \emph{what} reasoning is required but also about \emph{which framings} of that reasoning are accessible. 
QueST inherits this signal without any explicit framing conditioning.

\section{Failure Case Analysis}
\label{sec:failure}

We analyze cases where QueST fails to improve over the base model, including cases where test-time adaptation degrades performance.
These failures show that effective adaptation depends not only on generating auxiliary supervision, but also on whether that supervision captures the reasoning structure needed by the input query.
We identify two representative failure modes: surface-level matching and procedural retry without self-correction.

\begin{figure*}[t!]
    \centering
    \scalebox{1.00}{
    \includegraphics[width=1.0\textwidth]{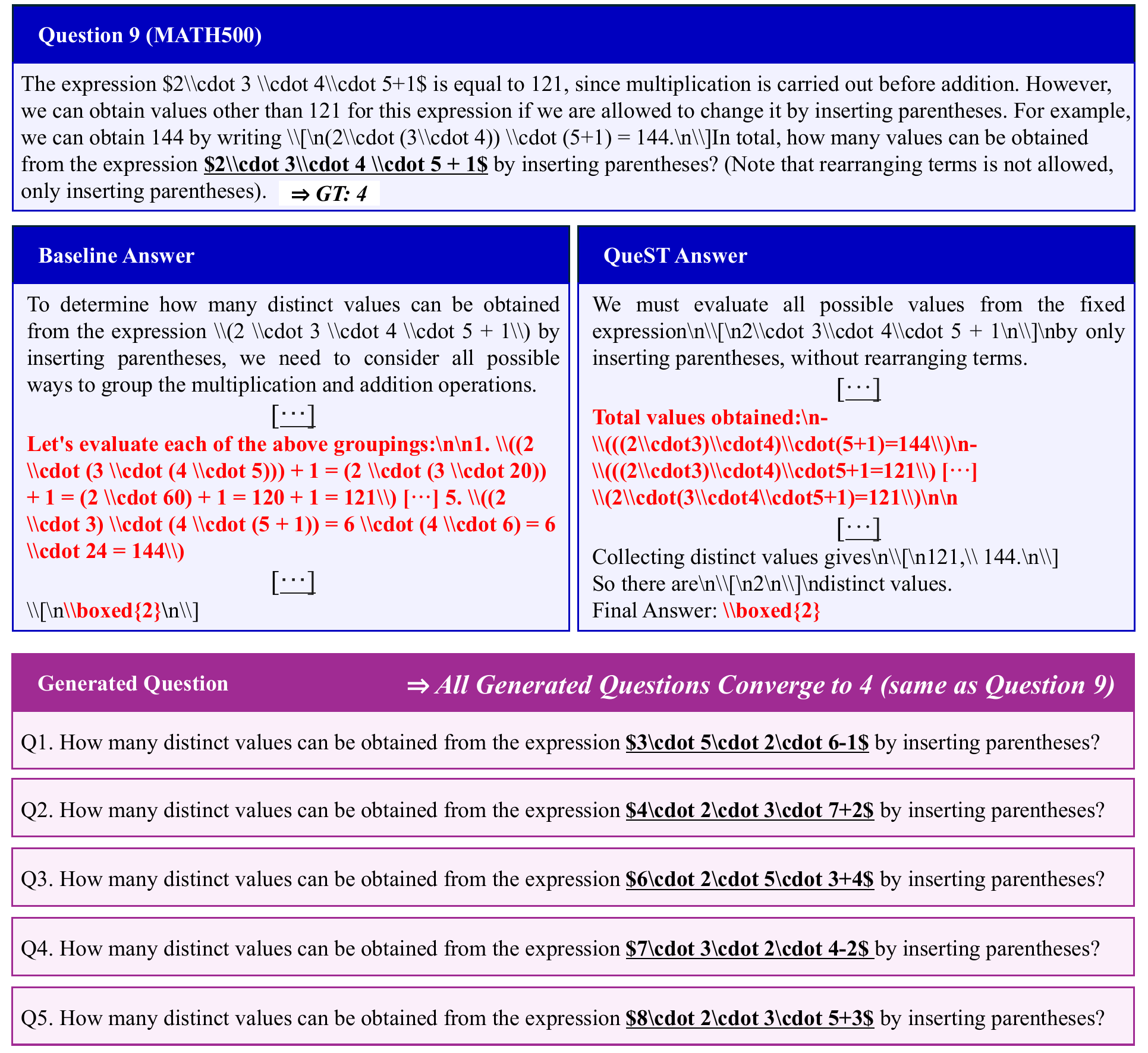}}
    \caption{Surface-level matching failure on a parenthesization-counting query from MATH500. The generated auxiliary problems share a near-identical syntactic template and coincidentally converge to the same answer as the original query. However, they fail to cover the enumeration logic required by the input, and the adapted model inherits the baseline's incorrect answer.}
    \label{fig:sample3} 
\end{figure*}

\paragraph{Surface-level Matching.}
The first failure mode arises when the generator matches the input query at the surface level, such as wording, numerical values, or syntactic form, without capturing the reasoning structure required to solve it. 
In this case, the auxiliary problems become numerical substitutions or trivial rephrasings of the input.
Although they appear relevant, they provide little supervision for the reasoning pattern that the original query requires.

Figure~\ref{fig:sample3} illustrates this on a parenthesization-counting problem from MATH500. 
Although the five auxiliary problems happen to share the input's ground-truth answer of $4$, they follow a nearly identical syntactic template.
As a result, they do not teach the enumeration logic required by the original query, and QueST inherits the baseline's incorrect answer $2$.

\begin{figure*}[t!]
    \centering
    \scalebox{1.00}{
    \includegraphics[width=1.0\textwidth]{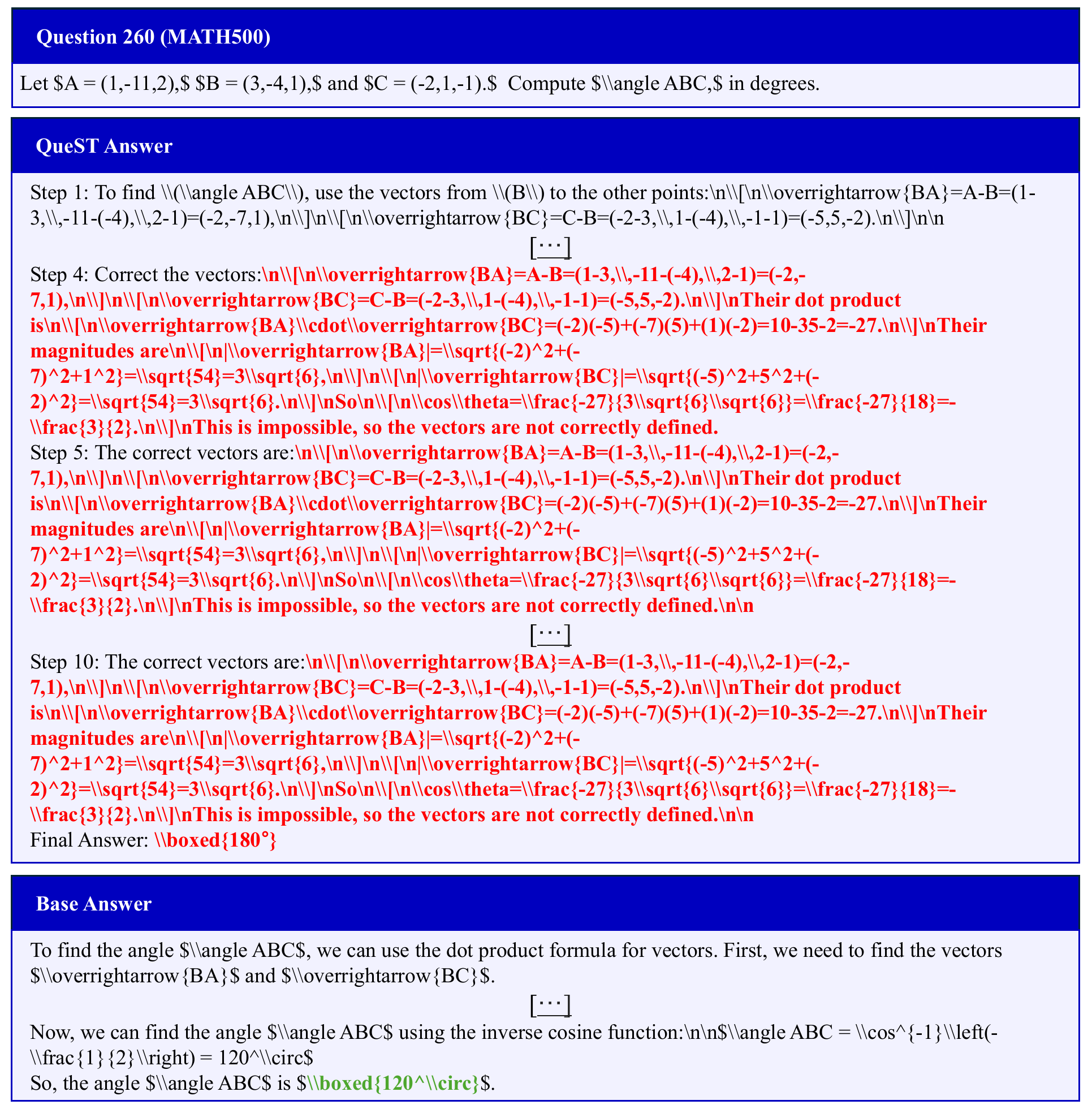}}
    \caption{Procedural retry failure on a vector-angle query. The adapted model repeatedly reproduces the same erroneous calculation block, even after recognizing that the intermediate cosine value is impossible. This leads to repeated procedural restarts rather than localized self-correction, and the final answer becomes incorrect.}
    \label{fig:sample4} 
\end{figure*}

\paragraph{Procedural Retry without Self-Correction.}

A second failure mode arises when the adapted model encounters its own computational error and attempts to recover. 
Because QueST's auxiliary supervision contains only successful solution traces, the adapted model learns to imitate the step-by-step procedural format of the supervision but is not exposed to examples of error diagnosis or self-correction.

When an error occurs at inference, the model can recognize that something is wrong, e.g., ``cos$\theta$ cannot be less than $-1$, so there must be a mistake.'' 
However, it lacks a learned mechanism to identify and revise the faulty step.
Instead, it opens a new numbered step and reproduces the same erroneous calculation.

Figure~\ref{fig:sample4} illustrates this behavior on a vector-angle problem: the adapted model recognizes the impossibility of its own intermediate result yet cannot localize the faulty step, and instead opens a new numbered step that reproduces the same erroneous block, resulting in a procedural retry rather than a substantive correction.

\paragraph{Implications.}
These two failure modes highlight two distinct limitations of the generated supervision: insufficient structural diversity and the absence of error-correction traces.
\textit{Surface-level matching} reflects a limitation in \emph{what the supervision contains}: the generated variants are aligned with the input at the surface level but do not cover the required reasoning structure.
In contrast, \textit{procedural retry} reflects a limitation in \emph{what the supervision omits}: the supervision contains successful solution traces but lacks examples of error diagnosis and recovery.
Both observations motivate future work on supervision generation that improves structural diversity and includes error-correction traces.

\section{Extended Ablation: Number of Generated Problems}
\label{sec:extended_n}
In the main paper, we report results for $N \in \{1, 3, 4, 5\}$ generated auxiliary problem--solution pairs (Table~\ref{tab:ablation}, Middle).
Here, we extend this ablation to larger values of $N \in \{6, 7, 8\}$ to examine whether additional supervision continues to provide gains beyond $N=5$ and to characterize the resulting accuracy--efficiency trade-off.

All experiments are conducted on Qwen3-4B-Base, evaluated on MATH500 with the same hyperparameters as the main experiments.
For each query, we generate $N$ auxiliary problem--solution pairs and perform test-time optimization on this expanded supervision set.

\begin{table}[t]
\centering
\renewcommand{\arraystretch}{1.15}
\caption{Effect of the number of generated problems on Qwen3-4B-Base evaluated on MATH500. Although accuracy continues to improve with larger $N$, we use $N=5$ in our main experiments to balance performance with the linearly growing test-time compute.}
\label{tab:extended_n}
\vspace{2mm} 
\begin{tabular}{c|c}
\toprule
\#\textbf{Problems} & \textbf{Acc} \\
\midrule
1 & 57.80 \\
3 & 63.00 \\
4 & 64.20 \\
5 & 68.60 \\
\midrule
6 & 69.20 \\
7 & 70.00 \\
8 & \textbf{70.40} \\
\bottomrule
\end{tabular}
\end{table}

Table~\ref{tab:extended_n} shows that QueST's performance continues to scale with $N$, improving from $68.60$ at $N{=}5$ to $70.40$ at $N{=}8$. 
The marginal gain per additional pair, however, diminishes from $+4.4$ ($N{:}4{\to}5$) to $+0.4$ ($N{:}7{\to}8$), indicating clear diminishing returns. 
Since per-query compute grows linearly with $N$, the choice of $N$ effectively controls the accuracy--efficiency trade-off. 
We adopt $N=5$ as the default operating point in our main experiments, as it captures most of the achievable accuracy at a substantially lower compute budget.

\section{Licenses for Existing Assets}
\label{sec:licenses}

We list the licenses and sources of all existing assets (models, datasets, and software libraries) used in this work. All assets are used for academic research and evaluation in accordance with their respective licenses, access conditions, and terms of use. For benchmarks based on AMC and AIME problems, we use publicly accessible past competition materials only for evaluation and do not redistribute the original problem sets.

\begin{table}[H]
\centering
\small
\setlength{\tabcolsep}{6pt}
\renewcommand{\arraystretch}{1.15}
\caption{Licenses and sources of all existing assets used in this work.}
\label{tab:licenses_all}
\vspace{2mm}
\resizebox{\linewidth}{!}{%
\begin{tabular}{lll}
\toprule
\textbf{Asset} & \textbf{License} & \textbf{Source} \\
\midrule
\multicolumn{3}{l}{\textit{Pre-trained Models}} \\
\midrule
Qwen3-4B~\cite{yang2025qwen3}               & Apache 2.0 & \url{https://huggingface.co/Qwen/Qwen3-4B} \\
Qwen3-8B~\cite{yang2025qwen3}               & Apache 2.0 & \url{https://huggingface.co/Qwen/Qwen3-8B} \\
Qwen3-4B-Base~\cite{yang2025qwen3}          & Apache 2.0 & \url{https://huggingface.co/Qwen/Qwen3-4B-Base} \\
Qwen3-8B-Base~\cite{yang2025qwen3}          & Apache 2.0 & \url{https://huggingface.co/Qwen/Qwen3-8B-Base} \\
OctoThinker-3B~\cite{wang2025octothinker}   & Llama 3.2 Community License & \url{https://huggingface.co/OctoThinker/OctoThinker-3B-Hybrid-Base} \\
OctoThinker-8B~\cite{wang2025octothinker}   & Llama 3.2 Community License & \url{https://huggingface.co/OctoThinker/OctoThinker-8B-Hybrid-Base}\\
\midrule
\multicolumn{3}{l}{\textit{Evaluation Benchmarks}} \\
\midrule
MATH500~\cite{hendrycks2021measuring, lightman2023let}  & MIT & \url{https://huggingface.co/datasets/HuggingFaceH4/MATH-500} \\
GSM8K~\cite{cobbe2021training}                          & MIT  & \url{https://huggingface.co/datasets/openai/gsm8k} \\
OlympiadBench~\cite{he2024olympiadbench}                & MIT & \url{https://huggingface.co/datasets/zwhe99/simplerl-OlympiadBench} \\
GPQA-Diamond~\cite{rein2023gpqa}                        & CC BY 4.0  & \url{https://huggingface.co/datasets/Idavidrein/gpqa} \\
Minerva~\cite{lewkowycz2022solving}                     & License not specified & \url{https://huggingface.co/datasets/zwhe99/simplerl-minerva-math} \\
AMC~\cite{maa_amc}                                      & MAA copyrighted (academic use) & \url{https://huggingface.co/datasets/zwhe99/amc23} \\
AIME24~\cite{maa_aime}                                  & MAA copyrighted (academic use) & \url{https://huggingface.co/datasets/HuggingFaceH4/aime_2024} \\
AIME25~\cite{maa_aime}                                  & MAA copyrighted (academic use) & \url{https://huggingface.co/datasets/yentinglin/aime_2025} \\
\midrule
\multicolumn{3}{l}{\textit{Software Libraries}} \\
\midrule
PyTorch                                                        & BSD-3-Clause & \url{https://github.com/pytorch/pytorch} \\
Hugging Face Transformers                                      & Apache 2.0   & \url{https://github.com/huggingface/transformers} \\
PEFT (LoRA)~\cite{hu2022lora}                                  & Apache 2.0   & \url{https://github.com/huggingface/peft} \\
TENT~\cite{wang2020tent}                                       & MIT          & \url{https://github.com/DequanWang/tent} \\
TLM~\cite{hu2025test}                                          & Apache 2.0   & \url{https://github.com/Fhujinwu/TLM} \\
\bottomrule
\end{tabular}
}
\end{table}


\newpage

\end{document}